\documentclass[runningheads]{llncs}
\usepackage[width=122mm,left=12mm,paperwidth=146mm,height=193mm,top=12mm,paperheight=217mm]{geometry}
\usepackage{graphicx}
\usepackage{amsmath,amssymb} 
\usepackage{color}
\usepackage[hidelinks]{hyperref}

\usepackage{rotating}
\usepackage{threeparttable}
\usepackage{array}
\usepackage{multirow}
\newcolumntype{C}[1]{>{\centering\let\newline\\\arraybackslash\hspace{0pt}}m{#1}}

\begin{document}
\title{Lightweight and Efficient\\Image Super-Resolution\\with Block State-based Recursive Network} 

\titlerunning{Super-Resolution with Block State-based Recursive Network}
%
\author{Jun-Ho Choi \and Jun-Hyuk Kim \and Manri Cheon \and Jong-Seok Lee}
%
\authorrunning{Choi et al.}
%

\institute{School of Integrated Technology, Yonsei University, Korea \\
	\email{\{idearibosome, junhyuk.kim, manri.cheon, jong-seok.lee\}@yonsei.ac.kr}}
\maketitle              
\begin{abstract}
Recently, several deep learning-based image super-resolution methods have been developed by stacking massive numbers of layers.
However, this leads too large model sizes and high computational complexities, thus some recursive parameter-sharing methods have been also proposed.
Nevertheless, their designs do not properly utilize the potential of the recursive operation.
In this paper, we propose a novel, lightweight, and efficient super-resolution method to maximize the usefulness of the recursive architecture, by introducing block state-based recursive network.
By taking advantage of utilizing the block state, the recursive part of our model can easily track the status of the current image features.
We show the benefits of the proposed method in terms of model size, speed, and efficiency.
In addition, we show that our method outperforms the other state-of-the-art methods.

\end{abstract}
\section{Introduction}

Single-image super-resolution is a task to obtain a high-resolution image from a given low-resolution image.
It is a kind of ill-posed problems since it has to estimate image details under the lack of spatial information.
Many researchers have proposed various approaches that can generate upscaled images having better quality than the simple interpolation methods such as nearest-neighbor, bilinear, and bicubic upscaling.

Recently, the emergence of deep learning techniques has flowed into the super-resolution field.
For example, Dong \textit{et al.} \cite{dong2014learning} proposed the super-resolution convolutional neural network (SRCNN) model, which showed much improved performance in comparison to the previous approaches.
Lim \textit{et al.} \cite{lim2017enhanced} suggested the enhanced deep super-resolution (EDSR) model, which employs residual connections and various optimization techniques.

Many recent deep learning-based super-resolution methods tend to stack much more numbers of layers to obtain better upscaled images, but this dramatically increases the number of involved model parameters.
For instance, the EDSR model requires about 43M parameters, which are at least 400 times more than those of the SRCNN model.
To deal with this, recursive approaches that use some parameters repeatedly have been proposed, including deeply-recursive convolutional network (DRCN) \cite{kim2016deeply}, deep recursive residual network (DRRN) \cite{tai2017image}, and dual-state recurrent network (DSRN) \cite{han2018image}.

The recursive super-resolution methods can be regarded as kinds of recurrent neural networks (RNNs) \cite{han2018image}.
RNNs have been usually employed when sequential relation of the data is significant, such as language modeling \cite{vinyals2015show} and human activity recognition \cite{choi2017impact}.
The beauty of RNNs comes from their two-fold structure: the recurrent unit handles not only the current input data but also the previously processed features.
Since the previously processed features contain historical information, RNNs can deal with sequential dependency of the inputs properly.

However, two characteristics of the existing recursive super-resolution methods hinder them from fully exploiting the usefulness of the RNNs.
First, there are no intermediate inputs and only the previously processed features are provided to the recurrent unit.
Second, the final output of the recurrent unit is directly used to obtain the final upscaled image.
In this situation, the output of the recurrent unit has to contain not only the super-resolved features, but also the historical information that is not useful in the non-recursive post-processing part.

\begin{figure}[t]
	\begin{center}
		\centering
		\includegraphics[width=0.65\linewidth]{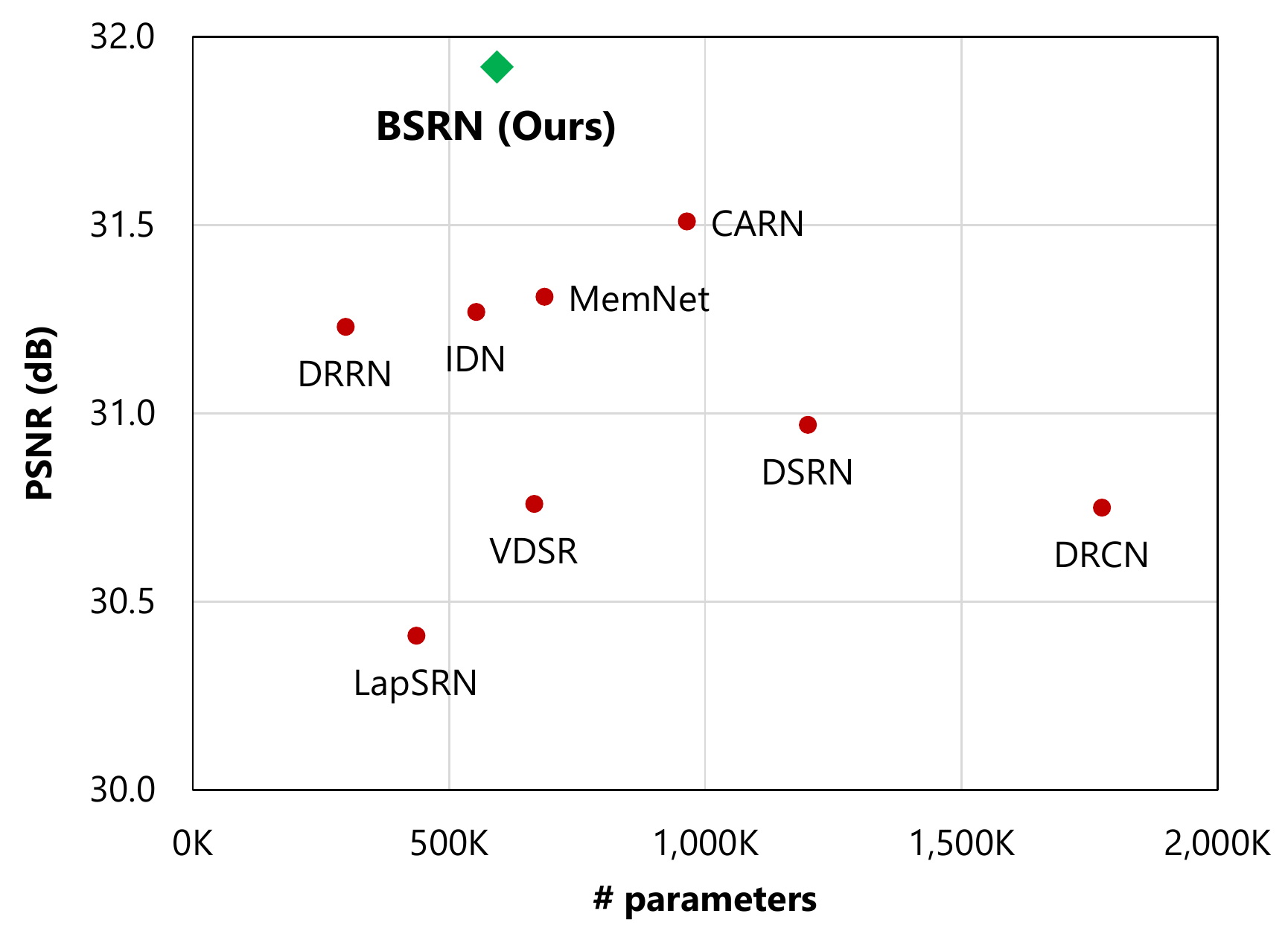}
	\end{center}
	\caption{Number of parameters and peak signal-to-noise ratio (PSNR) values of the state-of-the-art and the proposed methods for an upscaling factor of 2 on the Urban100 dataset \cite{huang2015single}.}
	\label{fig:params_psnr_graph}
\end{figure}

To alleviate this problem, we propose a novel super-resolution method using block state-based recursive network (BSRN).
Our method employs so-called ``block state'' along with the input features in the recursive part, which is a separate information storage to keep historical features.
Thanks to the elaborate design, our method achieves various benefits on top of the previous recursive super-resolution methods, in terms of image quality, lightness, speed, and efficiency.
As shown in \figurename~\ref{fig:params_psnr_graph}, our method achieves the best performance in terms of image quality, while the model complexity is significantly reduced.
In addition, the BSRN model can generate the super-resolved images in a progressive manner, which is useful for real-world applications such as progressive image loading.

The rest of the paper is organized as follows.
First, we discuss the related work in Section~\ref{sec:related_work}.
Then, the overall structure of the proposed method is explained in Section~\ref{sec:method}.
We present several experiments for in-depth analysis of our method in Section~\ref{sec:experiments}, including examining effectiveness of the newly introduced recursive structure and comparison with the other state-of-the-art methods.
Finally, we conclude our work in Section~\ref{sec:conclusion}.

\section{Related Work}
\label{sec:related_work}

Before deep learning has emerged, feature extraction-based methods have been widely used for super-resolution, such as sparse representation-based \cite{yang2011multitask} and Bayes forest-based \cite{salvador2015naive} approaches.
This trend has changed since deep learning showed significantly better performance in image classification tasks \cite{krizhevsky2012imagenet}.
Dong \textit{et al.} \cite{dong2014learning} pioneered the deep learning-based super-resolution by introducing SRCNN, which enhances the interpolated image via three convolutional layers.
Kim \textit{et al.} \cite{kim2016accurate} proposed very deep super-resolution (VDSR), which stacks 20 convolutional layers to improve the performance.
Lim \textit{et al.} \cite{lim2017enhanced} suggested the EDSR model, which employs more than 64 convolutional layers.
These methods share the basic empirical rule of deep learning: deeper and larger models can achieve better performance \cite{montufar2014number}.

As we addressed in the introduction, super-resolution methods sharing model parameters have been proposed.
DRCN introduced by Kim \textit{et al.} \cite{kim2016deeply} proves the effectiveness of parameter sharing, which recursively applies the feature extraction layer for 16 times.
Tai \textit{et al.} \cite{tai2017image} proposed DRRN that employs residual network (ResNet) \cite{he2016deep} with sharing the model parameters.
They also proposed the memory network (MemNet) model \cite{tai2017memnet}, which contains groups of recursive parts called ``memory blocks'' with skip connections across them.
Han \textit{et al.} \cite{han2018image} considered DRCN and DRRN as the RNNs employing recurrent states, and proposed DSRN, which uses dual recurrent states.
Ahn \textit{et al.} \cite{ahn2018fast} developed the cascading residual network (CARN) model, which employs cascading residual blocks with sharing their model parameters.
Although these methods can be regarded as RNNs as Han \textit{et al.} mentioned \cite{han2018image}, none of them uses a separate state, which is used in only the recursive part and not in the non-recursive post-processing part.

Some researchers proposed super-resolution methods that do not rely on shared parameters but have small numbers of model parameters.
For example, Lai \textit{et al.} \cite{lai2017deep} introduced the Laplacian pyramid super-resolution network (LapSRN) method, which progressively upscales the input image by a factor of 2.
Hui \textit{et al.} \cite{hui2018fast} proposed the information distillation network (IDN) method, which employs long and short feature extraction paths to maximize the amount of extracted information from the given low-resolution image.
Along with the recursive super-resolution methods, the performance of these methods is also compared with that of our proposed method in Section~\ref{sec:compare_sota}.

We observe the following three common techniques from the previous work.
First, increasing the spatial resolution at the latter stage can reduce the computational complexity than upscaling at the initial stage \cite{ahn2018fast,choi2017deep,hui2018fast}.
Second, employing multiple residual connections is beneficial to obtain better upscaled images \cite{kim2018deep,tai2017image}.
Third, obtaining multiple upscaled images from the same super-resolution model and combining them into one provides better quality than acquiring a single image directly \cite{kim2016deeply,lim2017enhanced,tai2017memnet}.
Along with the newly introduced block state-based architecture, our proposed method is built with considering the aforementioned empirical knowledge.

\section{Proposed Method}
\label{sec:method}

\begin{figure*}[t]
	\begin{center}
		\centering
		\includegraphics[width=0.95\linewidth]{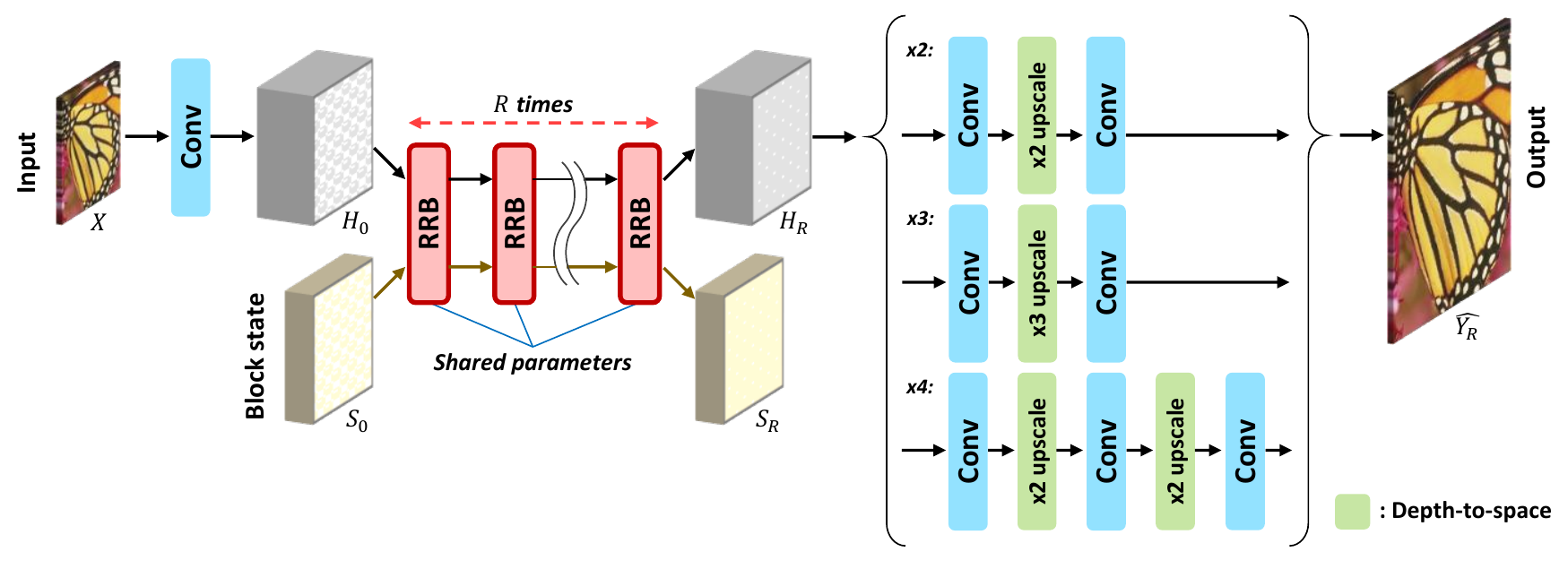}
	\end{center}
	\caption{Overall structure of the proposed BSRN model.}
	\label{fig:structure_main}
\end{figure*}

\begin{figure}[t]
	\begin{center}
		\centering
		\includegraphics[width=0.45\linewidth]{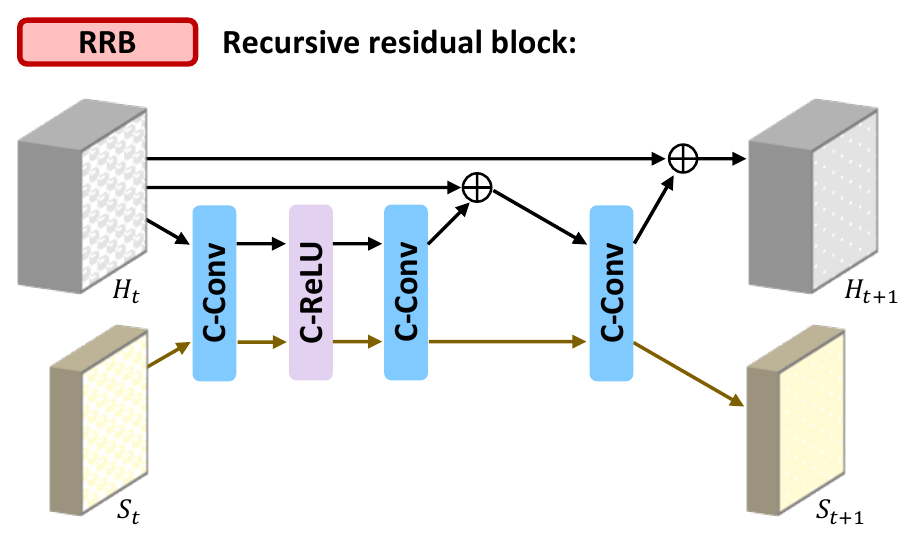}
	\end{center}
	\caption{Structure of a recursive residual block (RRB).}
	\label{fig:structure_rrb}
\end{figure}

In this section, we present how our super-resolution model works in detail.
As similar to the existing super-resolution methods, our BSRN model can be divided into three parts: initial feature extraction, feature processing in a recursive manner, and upscaling.
\figurename~\ref{fig:structure_main} shows the overall structure of our method.
As shown in the figure, the main objective of the super-resolution task is to obtain an image $\widehat{Y}$, which is upscaled from a given low-resolution image $X$, where we want $\widehat{Y}$ to be the same as the ground-truth image $Y$.
Briefly, the initial features are extracted from the given input image.
Then, the extracted features are further processed via a recursive residual block (RRB, \figurename~\ref{fig:structure_rrb}), which is employed multiple times with the same parameters.
The final image is obtained from the upscaling module.

\subsection{Initial feature extractor}

The BSRN model takes a low-resolution input image ${X}\in\mathbb{R}^{{w}\times{h}\times{3}}$ consisting of three channels of the RGB color space, where ${w}\times{h}$ is the resolution of the image.
Before we recursively process it, a convolutional layer extracts the initial features of the image, which can be represented as
\begin{equation}
{H}_{0} = {W}_{I} \ast {X} + {b}_{I}
\end{equation}
where ${W}_{I}\in\mathbb{R}^{{3}\times{3}\times{3}\times{c}}$ and ${b}_{I}\in\mathbb{R}^{c}$ are the weight and bias matrices, respectively, and the operator $\ast$ denotes the convolution operation.
A variable $c$ determines the number of convolutional channels, thus the last dimension of ${H}_{0}$ is $c$.

\subsection{Recursive residual block}
\label{sec:rrb}

Starting from the initial features ${H}_{0}$, our model performs the recursive operations in the shared part named ``recursive residual block (RRB),'' which is shown in \figurename~\ref{fig:structure_rrb}.
The RRB takes two matrices as inputs at a given iteration $t$: the feature matrix ${H}_{t}$ that has been processed at previous iterations from the original input image and an additional matrix ${S}_{t}\in\mathbb{R}^{{w}\times{h}\times{s}}$ called ``block state,'' where $s$ determines the feature dimension of ${S}_{t}$.
As shown in \figurename~\ref{fig:structure_main}, the block state matrix is not derived from the input image features.
Instead, the initial block state matrix ${S}_{0}$ is initialized by zero values.
Note that ${S}_{t}$ and ${H}_{t}$ have the same spatial dimension but different feature dimensions.

A RRB consists of three concatenated convolution (C-Conv) layers and one concatenated rectified linear unit (C-ReLU) layer.
A C-Conv layer first concatenates two input matrices along the last dimension, performs a convolutional operation, and splits the result into two output matrices having the sizes of the input matrices.
In other words, when ${H}_{t}$ and ${S}_{t}$ are given, a C-Conv layer concatenates them (i.e., $[{H}_{t}, {S}_{t}]$), applies convolution as
\begin{equation}
[{H}_{t}', {S}_{t}'] = {W}_{C}\ast [{H}_{t}, {S}_{t}] + {b}_{C}
\end{equation}
and splits them into ${H}_{t}'\in\mathbb{R}^{{w}\times{h}\times{c}}$ and ${S}_{t}'\in\mathbb{R}^{{w}\times{h}\times{s}}$, where ${W}_{C}\in\mathbb{R}^{3\times3\times{(c+s)}\times{(c+s)}}$ and ${b}_{C}\in\mathbb{R}^{(c+s)}$ are the weight and bias matrices, respectively.
A C-ReLU layer performs element-wise ReLU operations for the two inputs.
In addition, two residual connections are involved for better performance as in the previous work \cite{kim2016accurate,kim2018deep}.
After processing ${H}_{t}$ and ${S}_{t}$ with three C-Conv layers, one C-ReLU layer, and two residual connections for the ${H}_{t}$ part, the RRB outputs ${H}_{t+1}$ and ${S}_{t+1}$, which then serve as the inputs of the same RRB for the next recursion.
This recursive process is performed $R$ times, which produces ${H}_{R}$ and ${S}_{R}$.

There are two ways of configuring the BSRN model to get better performance: increasing the number of convolutional channels $c$ and increasing the feature dimension of the block state $s$.
When $c$ increases, the number of model parameters increases across all parts of the model, including the initial feature extraction, RRB, and upscaling parts.
On the other hand, increased $s$ affects the number of model parameters only in the RRB part because the block state is involved only in RRB.
Therefore, employing the block state is more beneficial to make the model compact than using a larger number of the convolutional channels.

In addition, because the block state can serve as a ``memory,'' the RRB can keep track of the status of the current image features over the recursive operations.
When the block state does not exist, it is hard to track the current status because it has to be latently written on the image features (i.e., ${H}_{t}$).
It may lead to quality degradation of upscaled images, since both the image features and the current status are inputted to the upscaling part.
We investigate the effectiveness of employing the block state in Section~\ref{sec:block_state_benefits}.

\subsection{Upscaling}

Finally, the BSRN model upscales the processed feature matrix ${H}_{R}$ to generate an upscaled image $\widehat{{Y}_{R}}$.
In particular, we use the depth-to-space operation as in the previous super-resolution models \cite{ahn2018fast,choi2017deep}, which is also known as sub-pixel convolution \cite{shi2016real}.
For instance, in the upscaling part by a factor of 2, the first convolutional layer outputs the processed matrix having a size of ${w}\times{h}\times{4c}$, the depth-to-space operator modifies the shape of the matrix to ${2w}\times{2h}\times{c}$, and the last convolutional layer outputs the final upscaled image having a shape of ${2w}\times{2h}\times{3}$.
Note that the block state ${S}_{t}$ is not used in the upscaling part.

Our model can generate upscaled images not only from the final processed feature matrix ${H}_{R}$ but also from the intermediate feature matrices ${H}_{t},~\forall {t}\in\{1, ..., R-1\}$.
Therefore, with our model, it is possible to generate the upscaled images in a progressive manner.
In addition, it is known that combining multiple outputs can improve the quality of the super-resolved images \cite{han2018image,kim2016deeply}.
Thus, we adopt a similar approach to obtain the final upscaled image $\widehat{Y}$ by combining the intermediate outputs via the weighted sum as:
\begin{equation}
\label{eq:y_hat}
\widehat{Y} = \frac{\sum_{t=1}^{R/r}{ {2}^{(rt-1)} \widehat{{Y}_{rt}} }}{\sum_{t=1}^{R/r}{ {2}^{(rt-1)} }}
\end{equation}
where $r$ is a so-called ``frequency control variable,'' which will be explained later.
The term ${2}^{(rt-1)}$ controls the amount of contribution of each intermediate output, where the later outputs contribute to $\widehat{Y}$ more than the earlier outputs.
This facilitates our model to generate intermediate upscaled images, which have progressively improved quality.

The variable $r$ in (\ref{eq:y_hat}) controls the frequency of the progressive upscaling.
For example, when $R=16$ and $r=4$, $\widehat{Y}$ is obtained from the weighted sum of $\widehat{{Y}_{4}}$, $\widehat{{Y}_{8}}$, $\widehat{{Y}_{12}}$, and $\widehat{{Y}_{16}}$.
Since a larger value of $r$ reduces the number of times to employ the upscaling part, it is beneficial to reduce the processing time for generating the final super-resolved image.
We discuss the influence of changing $r$ in Section~\ref{sec:r_role}.

\subsection{Loss function}

The loss function of our model is calculated from the weighted sum of the pixel-by-pixel L1 loss, i.e.,
\begin{equation}
\label{eq:loss}
L \big( \widehat{Y}, Y \big) = \frac{1}{w' \times h'} \sum_{x=1}^{w'} \sum_{y=1}^{h'} \left|  \widehat{Y}(x, y) - Y(x, y) \right|
\end{equation}
where $w' \times h'$ is the spatial resolution of $\widehat{Y}$ and $Y$, and $\widehat{Y}(x, y)$ and $Y(x, y)$ are the pixel values at $(x, y)$ of the upscaled and ground-truth images, respectively.

\section{Experiments}
\label{sec:experiments}

We conduct three experiments to investigate the advantages of the BSRN model.
First, we examine the effectiveness of employing the block state.
Second, we explore the role of the frequency control variable $r$.
Finally, we compare our models with the other state-of-the-art methods.

\subsection{Dataset and evaluation metrics}

We employ the DIV2K dataset \cite{agustsson2017ntire} for training the BSRN models, which is widely used for training the recent super-resolution models \cite{ahn2018fast,kim2018deep}.
For evaluating the performance of our models, we use four benchmark datasets, including Set5 \cite{bevilacqua2012low}, Set14 \cite{zeyde2010single}, BSD100 \cite{martin2001database}, and Urban100 \cite{huang2015single}.

We employ peak signal-to-noise ratio (PSNR) and structural similarity (SSIM) \cite{wang2004image} for measuring quality of the upscaled images.
As in the previous work \cite{ahn2018fast,hui2018fast}, both metrics are calculated on the Y channel of the YCbCr channels converted from the RGB channels.

\subsection{Training details}

We build both single-scale ($\times$4) and multi-scale ($\times$2, $\times$3, and $\times$4) BSRN models.
The single-scale models are used to find out the benefits of the block state and frequency control variable, and the multi-scale model is used to evaluate the performance of our model in comparison to the other super-resolution methods across different scales.
The number of the recursive operations $R$ and the frequency control variable $r$ are set to 16 and 1, respectively.

We implement the training and evaluation code of the BSRN model on the TensorFlow framework \cite{abadi2016tensorflow}\footnote{The code is available at \url{https://github.com/idearibosome/tf-bsrn-sr}.}.
For each training step, eight image patches are randomly cropped from the training images.
A cropping size of 32$\times$32 pixels is used for training the single-scale BSRN model and 48$\times$48 pixels is used for the multi-scale BSRN model.
For data augmentation, the image patches are then randomly flipped and rotated.
For the multi-scale BSRN model, one of the upscaling paths (i.e., $\times2$, $\times3$, and $\times4$) is randomly selected for every training step.
The super-resolved images are obtained from our model by feeding the image patches.
Then, the loss is calculated using (\ref{eq:loss}) and the Adam optimization method \cite{kingma2014adam} with $\beta_{1}=0.9$, $\beta_{2}=0.999$, and $\hat{\epsilon}={10}^{-8}$ is used to update the model parameters.
To prevent the vanishing or exploding gradients problem \cite{bengio1994learning}, we employ the L2 norm-based gradient clipping method, which clips each gradient so as to fit its L2 norm within $[-\theta, \theta]$.
In this study, we set $\theta=5$.
The initial learning rate is set to ${10}^{-4}$ and reduced by a half at every ${2}\times{10}^{5}$ training steps.
A total of ${1.0}\times{10}^{6}$ and ${1.5}\times{10}^{6}$ steps are executed for training the single-scale and multi-scale BSRN models, respectively.

\subsection{Benefits of employing the block state}
\label{sec:block_state_benefits}

\begin{figure}[t!]
	\begin{center}
		\centering
		\includegraphics[width=0.65\linewidth]{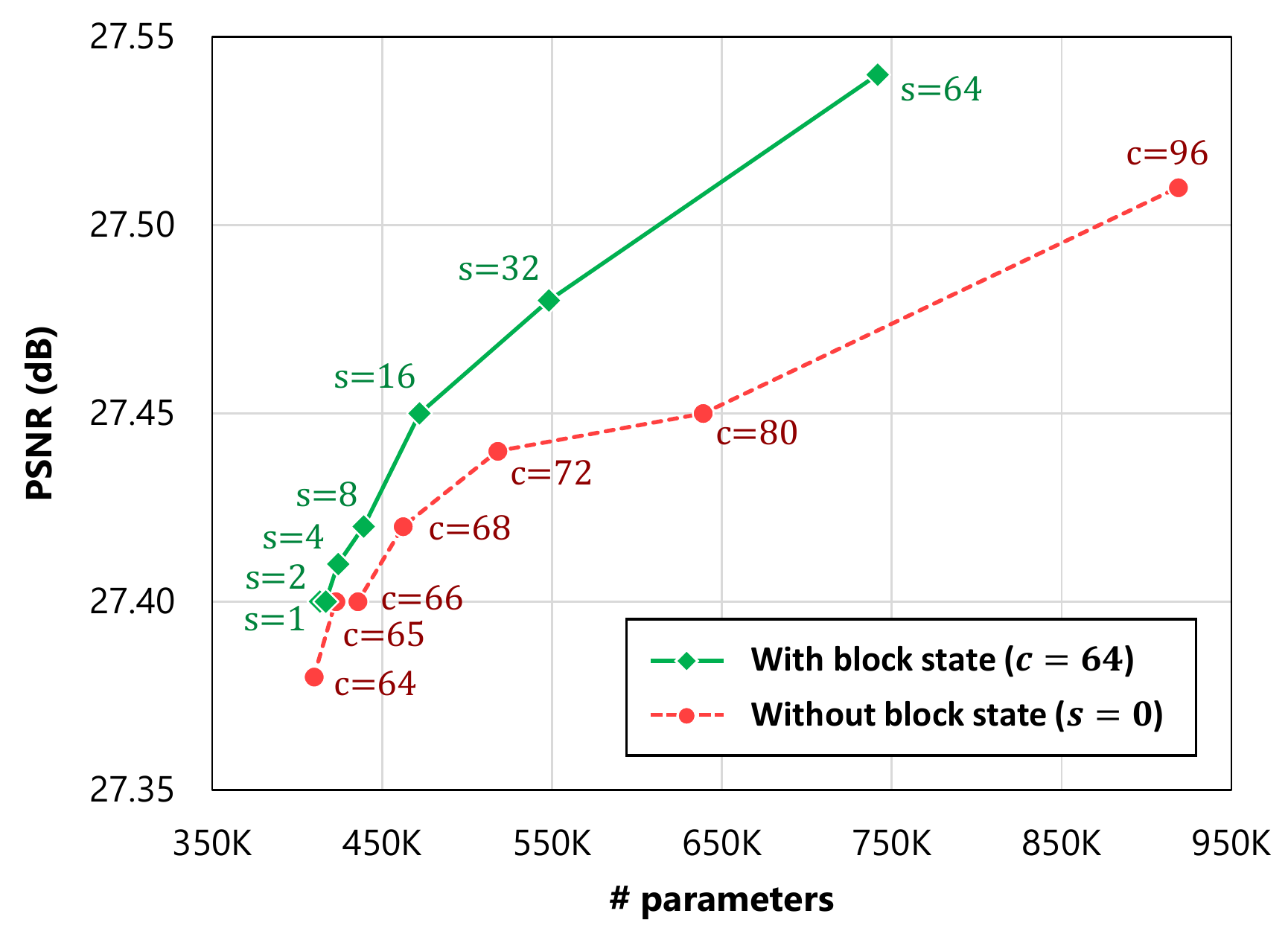}
	\end{center}
	\caption{Number of parameters and PSNR values of the $\times$4-scale BSRN models with and without the block state for the BSD100 dataset \cite{martin2001database}.}
	\label{fig:block_state_params_psnr}
\end{figure}

As explained in Section~\ref{sec:rrb}, the BSRN model can be trained with various numbers of the convolutional channels (i.e., $c$) and the block state channels (i.e., $s$).
Here, we investigate the effectiveness of employing the block state by comparing the single-scale BSRN models having an upscaling factor of 4, which are trained with and without using the block state.
For the models with the block state, the number of the convolutional channels $c$ is fixed to 64 and the number of the block state channels $s$ is changed from 1 to 64.
For the models without the block state, on the other hand, $s$ is fixed to 0 and $c$ is changed from 64 to 96.
All models are tested with $r=1$.

\begin{figure*}[t!]
	\begin{center}
		\centering
		\begin{minipage}[b]{0.995\linewidth}
			\centering
			\centerline{\includegraphics[width=1.0\linewidth]{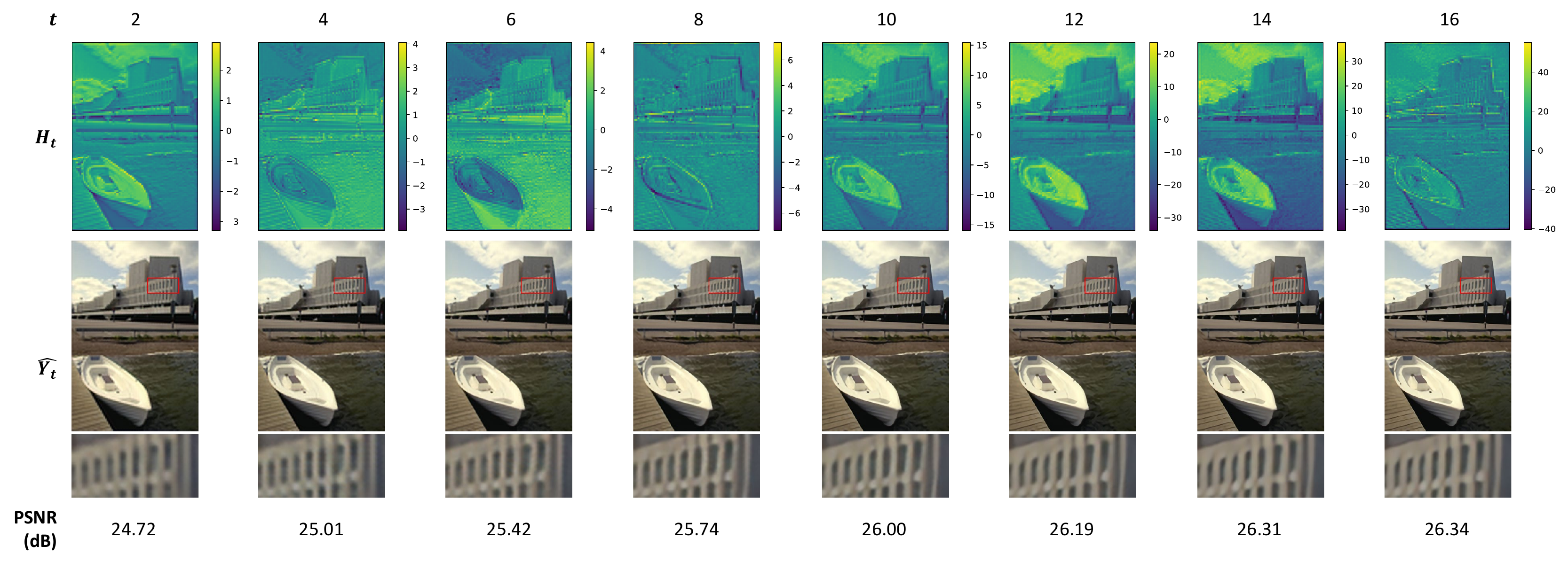}}
			\centerline{(a)}
		\end{minipage}
		\begin{minipage}[b]{0.995\linewidth}
			\centering
			\centerline{\includegraphics[width=1.0\linewidth]{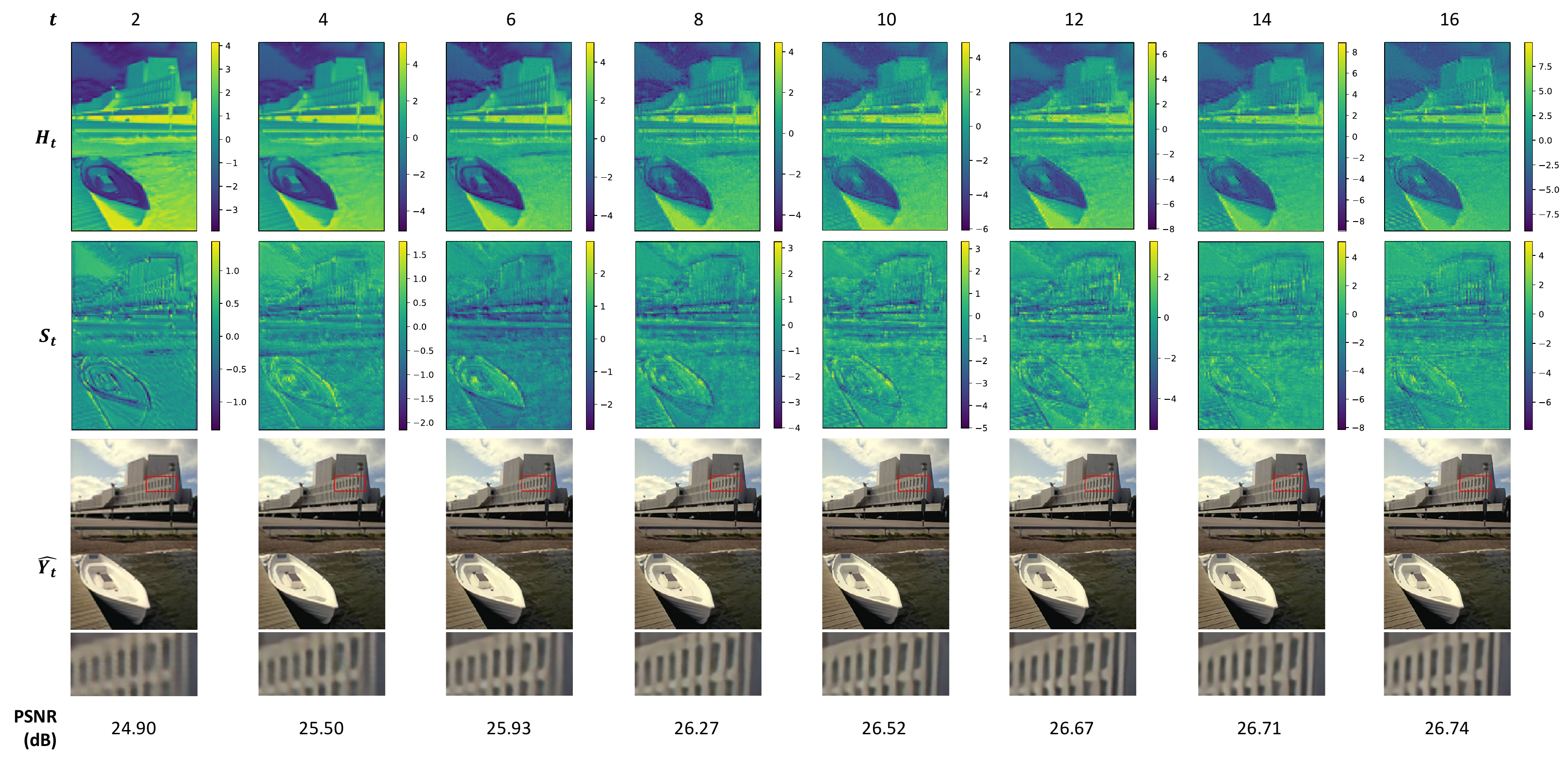}}
			\centerline{(b)}
		\end{minipage}
	\end{center}
	\caption{Intermediate features ${H}_{t}$, block states ${S}_{t}$, upscaled images $\widehat{{Y}_{t}}$, and PSNR values of the BSRN models from an image of the BSD100 dataset \cite{martin2001database}. Enlarged versions of the regions marked with red rectangles are also shown. (a) Without block state ($c=64, s=0$) (b) With block state ($c=64, s=64$)}
	\label{fig:state_features}
\end{figure*}

\figurename~\ref{fig:block_state_params_psnr} compares the performance of the trained BSRN models in terms of the number of parameters and the PSNR values measured for the BSD100 dataset \cite{martin2001database}.
Overall, both the models with and without having the block states have a tendency to show better performance as the feature dimension (and consequently the number of parameters) increases.
However, the BSRN models with the block state outperform the models without the block state, when the same numbers of parameters are used.
This strongly supports that differentiating the place to store historical information from that for the image features helps to improve the quality of the upscaled images.

We further examine changes of the activation patterns of the BSRN models over the recursive iterations.
\figurename~\ref{fig:state_features} shows ${H}_{t}$, ${S}_{t}$, and $\widehat{{Y}_{t}}$ of the BSRN models trained with $c=64, s=0$ and $c=64, s=64$, where the corresponding PSNR values are also reported.
The values of the intermediate features and block states are averaged along the last dimension.
Both the models with and without the block state generate the upscaled images with gradually improved quality in terms of the PSNR values over the iteration $t$.
However, the changes of the intermediate features are largely different.
When the block state is not employed (\figurename~\ref{fig:state_features} (a)), both the patterns of the activation and range of the values drastically change, even though the super-resolved images are not.
This implies that the RRB of the model without the block state has difficulty in generating progressively improved features and highly relies on the latter part (i.e., upscaling part) to generate good quality of the upscaled images.
On the other hand, employing the block state (\figurename~\ref{fig:state_features} (b)) results in much more stable activations of ${H}_{t}$ than the model without the block state.
Instead, the block states ${S}_{t}$ have major changes of the details, which provide historical information that can be used to produce gradually improved upscaled images over the iterations.
This confirms that our model properly utilizes the block state along with the intermediate output features, which leads to better performance.

\subsection{Role of frequency control variable ($r$)}
\label{sec:r_role}

\begin{table}[t]
	\scriptsize
	\begin{center}
		\begin{tabular}{c c c c}
			\noalign{\smallskip}
			\hline
			\noalign{\smallskip}
			~~~$r$~~~ & ~Processing time (s)~ & ~PSNR (dB)~ & ~SSIM~ \\
			\noalign{\smallskip}
			\hline
			\noalign{\smallskip}
			1 & 0.152 & 27.540 & 0.7341 \\
			2 & 0.094 & 27.540 & 0.7341 \\
			4 & 0.059 & 27.539 & 0.7341 \\
			8 & 0.047 & 27.538 & 0.7341 \\
			16 & 0.027 & 27.538 & 0.7341 \\
			\noalign{\smallskip}
			\hline
			\noalign{\smallskip}
		\end{tabular}
	\end{center}
	\caption{Performance comparison of the $\times$4-scale BSRN model tested with different values of $r$ in terms of the average processing time to obtain $\widehat{Y}$, PSNR, and SSIM for the BSD100 dataset \cite{martin2001database}.}
	\label{table:frequency_control_speed}
\end{table}

Our model can be configured with the frequency of progressive outputs via $r$, along with the number of recursive iterations $R$.
While $R$ determines how many times the RRB is used to generate the final upscaled image, $r$ determines how many intermediate images are obtained from the model to generate the final image (i.e., how many times the upscaling part is employed), which is $R/r$.
Note that both $R$ and $r$ do not affect the number of model parameters.

In our proposed model, the upscaling part spends most of the computation time due to its increased number of the convolutional filters for the depth-to-space operation and increased spatial resolution after the depth-to-space operation.
To verify this, we examine the BSRN model trained with $c=64$ and $s=64$ by testing with different values of $r$ and compare their efficiency in terms of speed and quality of the upscaled images (i.e., PSNR and SSIM).

Table~\ref{table:frequency_control_speed} shows the average processing time spent on upscaling an image by a factor of 4, PSNR values, and SSIM values for the BSD100 dataset \cite{martin2001database} for various values of $r$.
The processing time is measured on a NVIDIA GeForce GTX 1080 GPU.
As expected, the processing time largely decreases when $r$ increases.
For example, the BSRN model tested with $r=16$ requires more than 5 times less processing time than the model tested with $r=1$.
Nevertheless, the PSNR value decreases by only 0.002 dB and SSIM value even remains the same.
This confirms that increasing $r$ significantly increases the processing speed with only negligible quality degradation.
In addition, the experimental result implies that our proposed model has a capability of real-time processing.
For example, when $r=16$, our model can upscale more than 30 images per second, which is a common frame rate of videos.

\subsection{Comparison with the other methods}
\label{sec:compare_sota}

\begin{table*}[t!]
	\scriptsize
	\begin{center}
		\begin{tabular}{c l r l l l l}
			\noalign{\smallskip}
			\hline
			\noalign{\smallskip}
			\multirow{2}{*}{~Scale~} & \multirow{2}{*}{Method} & \multirow{2}{*}{\# params~} & Set5 & Set14 & BSD100 & Urban100 \\
			& & & PSNR / SSIM~ & PSNR / SSIM~ & PSNR / SSIM~ & PSNR / SSIM~ \\
			\noalign{\smallskip}
			\hline
			\noalign{\smallskip}
			\multirow{9}{*}{$\times$2} & VDSR \cite{kim2016accurate} & 666K~ & 37.53 / 0.9587 & 33.03 / 0.9124 & 31.90 / 0.8960 & 30.76 / 0.9140 \\
			& DRCN \cite{kim2016deeply} & 1,774K~ & 37.63 / 0.9588 & 33.04 / 0.9118 & 31.85 / 0.8942 & 30.75 / 0.9133 \\
			& LapSRN \cite{lai2017deep} & 436K~ & 37.52 / 0.959~~ & 33.08 / 0.913~~ & 31.80 / 0.895~~ & 30.41 / 0.910~~ \\
			& DRRN \cite{tai2017image} & 298K~ & 37.74 / 0.9591 & 33.23 / 0.9136 & 32.05 / 0.8973 & 31.23 / 0.9188 \\
			& MemNet \cite{tai2017memnet} & 686K~ & \textcolor{blue}{37.78} / \textcolor{blue}{0.9597} & 33.23 / 0.9142 & 32.08 / 0.8978 & 31.31 / 0.9195 \\
			& DSRN \cite{han2018image} & $\sim$1,200K~ & 37.66 / 0.959~~ & 33.15 / 0.913~~ & \textcolor{blue}{32.10} / 0.897~~ & 30.97 / 0.916~~ \\
			& IDN \cite{hui2018fast} & 553K~ & \textcolor{red}{37.83} / \textcolor{red}{0.9600} & 33.30 / 0.9148 & 32.08 / \textcolor{red}{0.8985} & 31.27 / 0.9196 \\
			& CARN \cite{ahn2018fast} & 964K~ & 37.76 / 0.9590 & \textcolor{red}{33.52} / \textcolor{red}{0.9166} & 32.09 / 0.8978 & \textcolor{blue}{31.51} / \textcolor{red}{0.9312} \\
			& \textbf{BSRN (Ours)} & 594K~ & \textcolor{blue}{37.78} / 0.9591 & \textcolor{blue}{33.43} / \textcolor{blue}{0.9155} & \textcolor{red}{32.11} / \textcolor{blue}{0.8983} & \textcolor{red}{31.92} / \textcolor{blue}{0.9261} \\
			\noalign{\smallskip}
			\hline
			\noalign{\smallskip}
			\multirow{8}{*}{$\times$3} & VDSR \cite{kim2016accurate} & 666K~ & 33.66 / 0.9213 & 29.77 / 0.8314 & 28.82 / 0.7976 & 27.14 / 0.8279 \\
			& DRCN \cite{kim2016deeply} & 1,774K~ & 33.82 / 0.9226 & 29.76 / 0.8311 & 28.80 / 0.7963 & 27.15 / 0.8276 \\
			& DRRN \cite{tai2017image} & 298K~ & 34.03 / 0.9244 & 29.96 / 0.8349 & 28.95 / 0.8004 & 27.53 / 0.8378 \\
			& MemNet \cite{tai2017memnet} & 686K~ & 34.09 / 0.9248 & 30.00 / 0.8350 & 28.96 / 0.8001 & \textcolor{blue}{27.56} / 0.8376 \\
			& DSRN \cite{han2018image} & $\sim$1,200K~ & 33.88 / 0.922~~ & \textcolor{blue}{30.26} / 0.837~~ & 28.81 / 0.797~~ & 27.16 / 0.828~~ \\
			& IDN \cite{hui2018fast} & 553K~ & 34.11 / 0.9253 & 29.99 / 0.8354 & 28.95 / 0.8013 & 27.42 / 0.8359 \\
			& CARN \cite{ahn2018fast} & 1,149K~ & \textcolor{blue}{34.29} / \textcolor{red}{0.9255} & \textcolor{red}{30.29} / \textcolor{red}{0.8407} & \textcolor{blue}{29.06} / \textcolor{blue}{0.8034} & 27.38 / \textcolor{blue}{0.8404} \\
			& \textbf{BSRN (Ours)} & 779K~ & \textcolor{red}{34.32} / \textcolor{red}{0.9255} & 30.25 / \textcolor{blue}{0.8404} & \textcolor{red}{29.07} / \textcolor{red}{0.8039} & \textcolor{red}{28.04} / \textcolor{red}{0.8497} \\
			\noalign{\smallskip}
			\hline
			\noalign{\smallskip}
			\multirow{9}{*}{$\times$4} & VDSR \cite{kim2016accurate} & 666K~ & 31.35 / 0.8838 & 28.01 / 0.7674 & 27.29 / 0.7251 & 25.18 / 0.7524 \\
			& DRCN \cite{kim2016deeply} & 1,774K~ & 31.53 / 0.8854 & 28.02 / 0.7670 & 27.23 / 0.7233 & 25.14 / 0.7510 \\
			& LapSRN \cite{lai2017deep} & 872K~ & 31.54 / 0.885~~ & 28.19 / 0.772~~ & 27.32 / 0.728~~ & 25.21 / 0.756~~ \\
			& DRRN \cite{tai2017image} & 298K~ & 31.68 / 0.8888 & 28.21 / 0.7720 & 27.38 / 0.7284 & 25.44 / 0.7638 \\
			& MemNet \cite{tai2017memnet} & 686K~ & 31.74 / 0.8893 & 28.26 / 0.7723 & 27.40 / 0.7281 & 25.50 / 0.7630 \\
			& DSRN \cite{han2018image} & $\sim$1,200K~ & 31.40 / 0.883~~ & 28.07 / 0.770~~ & 27.25 / 0.724~~ & 25.08 / 0.717~~ \\
			& IDN \cite{hui2018fast} & 553K~ & 31.82 / 0.8903 & 28.25 / 0.7730 & 27.41 / 0.7297 & 25.41 / 0.7632 \\
			& CARN \cite{ahn2018fast} & 1,112K~ & \textcolor{blue}{32.13} / \textcolor{red}{0.8937} & \textcolor{red}{28.60} / \textcolor{red}{0.7806} & \textcolor{red}{27.58} / \textcolor{blue}{0.7349} & \textcolor{red}{26.07} / \textcolor{red}{0.7837} \\
			& \textbf{BSRN (Ours)} & 742K~ & \textcolor{red}{32.14} / \textcolor{red}{0.8937} & \textcolor{blue}{28.56} / \textcolor{blue}{0.7803} & \textcolor{blue}{27.57} / \textcolor{red}{0.7353} & \textcolor{blue}{26.03} / \textcolor{blue}{0.7835} \\
			\noalign{\smallskip}
			\hline
			\noalign{\smallskip}
		\end{tabular}
	\end{center}
	\caption{Performance comparison of the state-of-the-art methods and our model evaluated on the Set5 \cite{bevilacqua2012low}, Set14 \cite{zeyde2010single}, BSD100 \cite{martin2001database}, and Urban100 \cite{huang2015single} datasets. Red and blue colors indicate the best and second best performance, respectively.}
	\label{table:performance_comparison}
\end{table*}

\begin{figure*}[t!]
\begin{center}
	\centering
	\begin{minipage}[b]{0.995\linewidth}
		\centering
		\centerline{\includegraphics[width=1.0\linewidth]{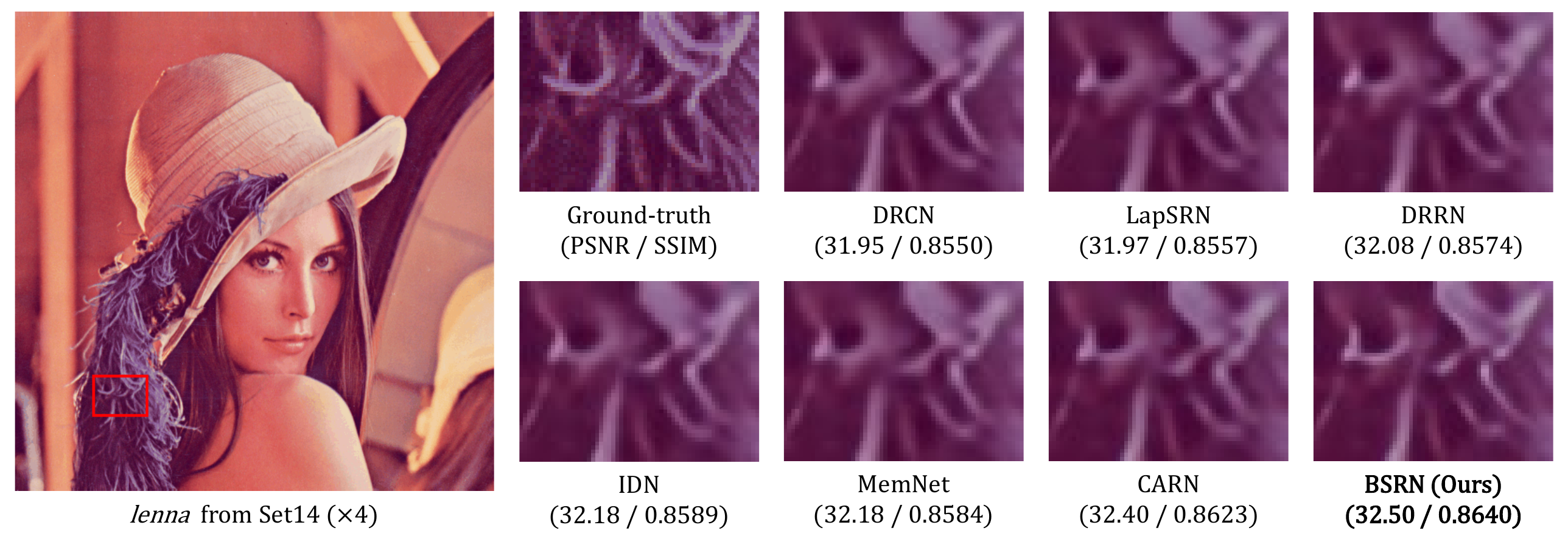}}
		\vspace{0.08in}
	\end{minipage}
	\begin{minipage}[b]{0.995\linewidth}
		\centering
		\centerline{\includegraphics[width=1.0\linewidth]{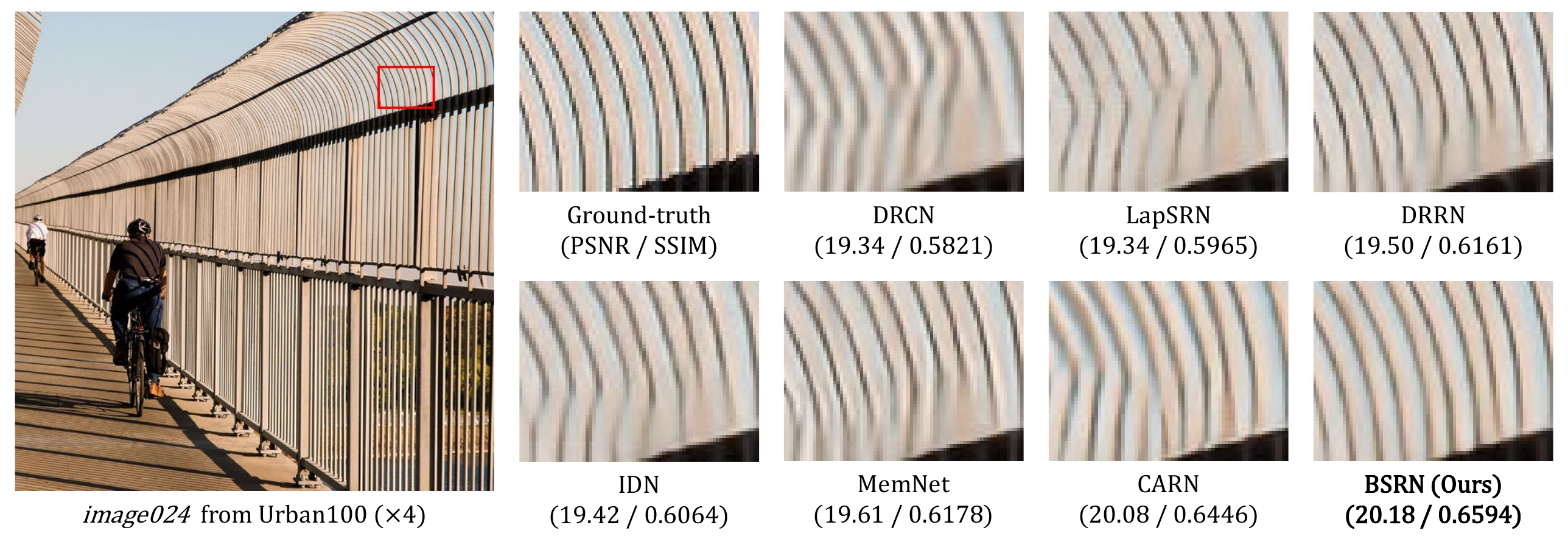}}
		\vspace{0.08in}
	\end{minipage}
	\begin{minipage}[b]{0.995\linewidth}
		\centering
		\centerline{\includegraphics[width=1.0\linewidth]{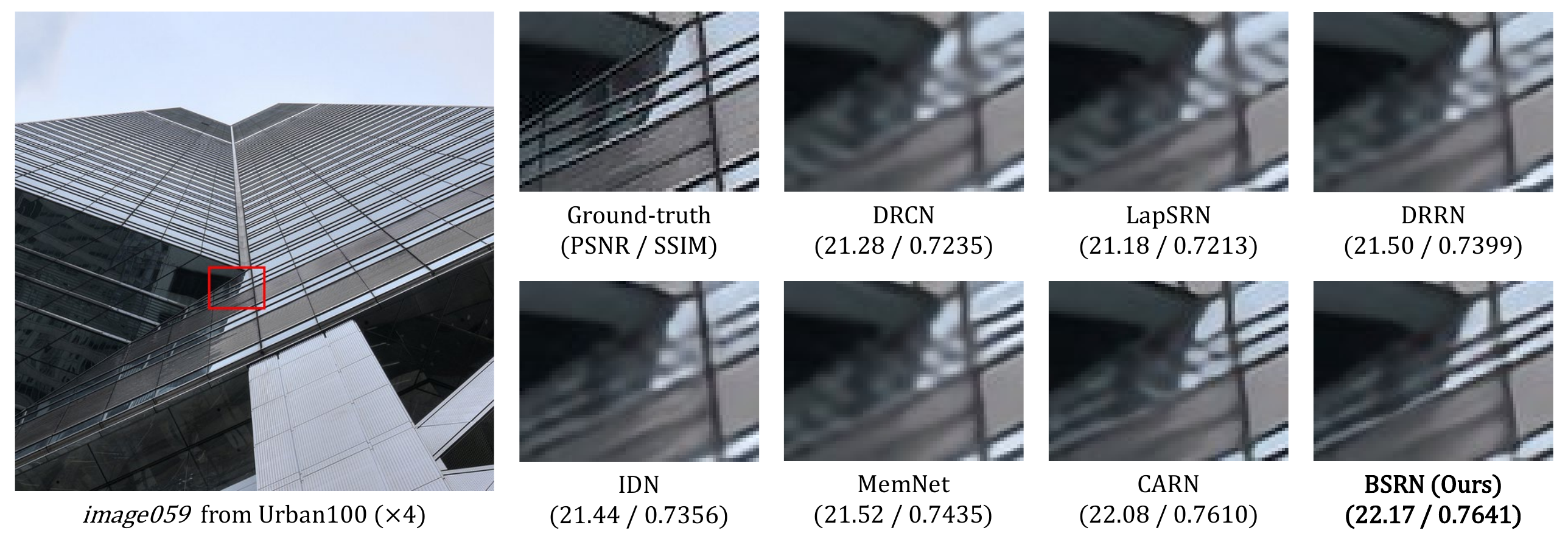}}
	\end{minipage}
\end{center}
\caption{Comparison of the upscaled images obtained by the BSRN model and the other state-of-the-art methods.}
\label{fig:sota_vs_bsrn}
\end{figure*}

Finally, we compare the performance of the multi-scale BSRN model with the other state-of-the-art super-resolution methods, including VDSR \cite{kim2016accurate}, DRCN \cite{kim2016deeply}, LapSRN \cite{lai2017deep}, DRRN \cite{tai2017image}, MemNet \cite{tai2017memnet}, DSRN \cite{han2018image}, IDN \cite{hui2018fast}, and CARN \cite{ahn2018fast}.
The DRCN, DRRN, MemNet, DSRN, and CARN models contain parameter-sharing parts.
The VDSR, LapSRN, and IDN methods are also included in the comparison, since they have been recently proposed and have similar numbers of model parameters to ours.

Table~\ref{table:performance_comparison} shows the performance of the state-of-the-art methods and ours in terms of the PSNR and SSIM values on the four benchmark datasets.
The number of model parameters required to obtain the super-resolved image with the given upscaling factor for each method is also provided.
First, the BSRN model outperforms the other methods that do not employ any recursive operations or parameter-sharing, including VDSR, LapSRN, and IDN.
For example, our method achieves a quality gain of 0.31 dB for a scale factor of 2 on the BSD100 dataset over the LapSRN model.
It confirms that recursive processing helps to obtain better super-resolved images with keeping the number of model parameters small enough.

In addition, our model employs much less numbers of parameters than DRCN, DSRN, and CARN.
For instance, the BSRN model uses up to 70\% less numbers of model parameters than the DRCN model.
Nevertheless, our proposed model outperforms DRCN and DSRN, and shows comparable performance to CARN.
In particular, BSRN shows almost the same performance as CARN despite the smaller model size.
This proves that the proposed method handles the image features better than the other state-of-the-art methods.

\figurename~\ref{fig:sota_vs_bsrn} provides a showcase of the images reconstructed by our proposed model and the other state-of-the-art methods.
The figure shows that the BSRN model is highly reliable in recovering textures from the low-resolution images.
For example, our method successfully upscales fine details of the structures in the Urban100 dataset, which results in clearer outputs, while the other methods produce highly blurred images or images containing large amounts of artifacts.
This confirms that the BSRN model produces images having visually nice super-resolved images.

\section{Conclusion}
\label{sec:conclusion}

In this paper, we introduced the BSRN model, which employs a novel way of recursive operation using the block state, for the super-resolution tasks.
We explained the benefits and efficiency of employing our model in terms of the number of model parameters, quality measures (i.e., PSNR and SSIM), and speed.
In addition, comparison with the other state-of-the-art methods also showed that our method can generate better quality of the upscaled images than the others.

%
%
%
\bibliographystyle{splncs04}
\bibliography{arxiv}

\begin{thebibliography}{10}
\providecommand{\url}[1]{\texttt{#1}}
\providecommand{\urlprefix}{URL }
\providecommand{\doi}[1]{https://doi.org/#1}

\bibitem{abadi2016tensorflow}
Abadi, M., Barham, P., Chen, J., Chen, Z., Davis, A., Dean, J., Devin, M.,
  Ghemawat, S., Irving, G., Isard, M., et~al.: {TensorFlow}: {A} system for
  large-scale machine learning. In: Proceedings of the USENIX Symposium on
  Operating Systems Design and Implementation. pp. 265--283 (2016)

\bibitem{agustsson2017ntire}
Agustsson, E., Timofte, R.: {NTIRE} 2017 challenge on single image
  super-resolution: {D}ataset and study. In: Proceedings of the IEEE Conference
  on Computer Vision and Pattern Recognition Workshops. pp. 126--135 (2017)

\bibitem{ahn2018fast}
Ahn, N., Kang, B., Sohn, K.A.: Fast, accurate, and lightweight super-resolution
  with cascading residual network. In: Proceedings of the European Conference
  on Computer Vision. pp. 252--268 (2018)

\bibitem{bengio1994learning}
Bengio, Y., Simard, P., Frasconi, P.: Learning long-term dependencies with
  gradient descent is difficult. IEEE Transactions on Neural Networks
  \textbf{5}(2),  157--166 (1994)

\bibitem{bevilacqua2012low}
Bevilacqua, M., Roumy, A., Guillemot, C., Alberi-Morel, M.L.: Low-complexity
  single-image super-resolution based on nonnegative neighbor embedding. In:
  Proceedings of the British Machine Vision Conference. pp. 1--10 (2012)

\bibitem{choi2017deep}
Choi, J.S., Kim, M.: A deep convolutional neural network with selection units
  for super-resolution. In: Proceedings of the IEEE Conference on Computer
  Vision and Pattern Recognition Workshops. pp. 1150--1156 (2017)

\bibitem{choi2017impact}
Choi, J.H., Choi, M., Choi, M.S., Lee, J.S.: Impact of three-dimensional video
  scalability on multi-view activity recognition using deep learning. In:
  Proceedings of the Thematic Workshops of ACM Conference on Multimedia. pp.
  135--143 (2017)

\bibitem{dong2014learning}
Dong, C., Loy, C.C., He, K., Tang, X.: Learning a deep convolutional network
  for image super-resolution. In: Proceedings of the European Conference on
  Computer Vision. pp. 184--199 (2014)

\bibitem{han2018image}
Han, W., Chang, S., Liu, D., Yu, M., Witbrock, M., Huang, T.S.: Image
  super-resolution via dual-state recurrent networks. In: Proceedings of the
  IEEE Conference on Computer Vision and Pattern Recognition. pp. 1654--1663
  (2018)

\bibitem{he2016deep}
He, K., Zhang, X., Ren, S., Sun, J.: Deep residual learning for image
  recognition. In: Proceedings of the IEEE Conference on Computer Vision and
  Pattern Recognition. pp. 770--778 (2016)

\bibitem{huang2015single}
Huang, J.B., Singh, A., Ahuja, N.: Single image super-resolution from
  transformed self-exemplars. In: Proceedings of the IEEE Conference on
  Computer Vision and Pattern Recognition. pp. 5197--5206 (2015)

\bibitem{hui2018fast}
Hui, Z., Wang, X., Gao, X.: Fast and accurate single image super-resolution via
  information distillation network. In: Proceedings of the IEEE Conference on
  Computer Vision and Pattern Recognition. pp. 723--731 (2018)

\bibitem{kim2016accurate}
Kim, J., Lee, J.K., Lee, K.M.: Accurate image super-resolution using very deep
  convolutional networks. In: Proceedings of the IEEE Conference on Computer
  Vision and Pattern Recognition. pp. 1646--1654 (2016)

\bibitem{kim2016deeply}
Kim, J., Lee, J.K., Lee, K.M.: Deeply-recursive convolutional network for image
  super-resolution. In: Proceedings of the IEEE Conference on Computer Vision
  and Pattern Recognition. pp. 1637--1645 (2016)

\bibitem{kim2018deep}
Kim, J.H., Lee, J.S.: Deep residual network with enhanced upscaling module for
  super-resolution. In: Proceedings of the IEEE Conference on Computer Vision
  and Pattern Recognition Workshops. pp. 800--808 (2018)

\bibitem{kingma2014adam}
Kingma, D.P., Ba, J.: Adam: {A} method for stochastic optimization. In:
  Proceedings of the International Conference on Learning Representations. pp.
  1--13 (2015)

\bibitem{krizhevsky2012imagenet}
Krizhevsky, A., Sutskever, I., Hinton, G.E.: {ImageNet} classification with
  deep convolutional neural networks. In: Proceedings of the Advances in Neural
  Information Processing Systems. pp. 1097--1105 (2012)

\bibitem{lai2017deep}
Lai, W.S., Huang, J.B., Ahuja, N., Yang, M.H.: Deep {L}aplacian pyramid
  networks for fast and accurate super-resolution. In: Proceedings of the IEEE
  Conference on Computer Vision and Pattern Recognition. pp. 624--632 (2017)

\bibitem{lim2017enhanced}
Lim, B., Son, S., Kim, H., Nah, S., Lee, K.M.: Enhanced deep residual networks
  for single image super-resolution. In: Proccedings of the IEEE Conference on
  Computer Vision and Pattern Recognition Workshops. pp. 136--144 (2017)

\bibitem{martin2001database}
Martin, D., Fowlkes, C., Tal, D., Malik, J.: A database of human segmented
  natural images and its application to evaluating segmentation algorithms and
  measuring ecological statistics. In: Proceedings of the IEEE International
  Conference on Computer Vision. pp. 416--423 (2001)

\bibitem{montufar2014number}
Montufar, G.F., Pascanu, R., Cho, K., Bengio, Y.: On the number of linear
  regions of deep neural networks. In: Proceedings of the Advances in Neural
  Information Processing Systems. pp. 2924--2932 (2014)

\bibitem{salvador2015naive}
Salvador, J., Perez-Pellitero, E.: Naive {B}ayes super-resolution forest. In:
  Proceedings of the IEEE International Conference on Computer Vision. pp.
  325--333 (2015)

\bibitem{shi2016real}
Shi, W., Caballero, J., Husz{\'a}r, F., Totz, J., Aitken, A.P., Bishop, R.,
  Rueckert, D., Wang, Z.: Real-time single image and video super-resolution
  using an efficient sub-pixel convolutional neural network. In: Proceedings of
  the IEEE Conference on Computer Vision and Pattern Recognition. pp.
  1874--1883 (2016)

\bibitem{tai2017image}
Tai, Y., Yang, J., Liu, X.: Image super-resolution via deep recursive residual
  network. In: Proceedings of the IEEE Conference on Computer Vision and
  Pattern Recognition. pp. 3147--3155 (2017)

\bibitem{tai2017memnet}
Tai, Y., Yang, J., Liu, X., Xu, C.: Mem{N}et: {A} persistent memory network for
  image restoration. In: Proceedings of the IEEE Conference on Computer Vision
  and Pattern Recognition. pp. 4539--4547 (2017)

\bibitem{vinyals2015show}
Vinyals, O., Toshev, A., Bengio, S., Erhan, D.: Show and tell: {A} neural image
  caption generator. In: Proceedings of the IEEE Conference on Computer Vision
  and Pattern Recognition. pp. 3156--3164 (2015)

\bibitem{wang2004image}
Wang, Z., Bovik, A.C., Sheikh, H.R., Simoncelli, E.P.: Image quality
  assessment: {F}rom error visibility to structural similarity. IEEE
  Transactions on Image Processing  \textbf{13}(4),  600--612 (2004)

\bibitem{yang2011multitask}
Yang, S., Liu, Z., Wang, M., Sun, F., Jiao, L.: Multitask dictionary learning
  and sparse representation based single-image super-resolution reconstruction.
  Neurocomputing  \textbf{74}(17),  3193--3203 (2011)

\bibitem{zeyde2010single}
Zeyde, R., Elad, M., Protter, M.: On single image scale-up using
  sparse-representations. In: Proceedings of the International Conference on
  Curves and Surfaces. pp. 711--730 (2010)

\end{thebibliography}
\end{document}